\documentclass{article}

\usepackage[accepted]{icml2026}

\usepackage{microtype}

\usepackage{graphicx}

\usepackage{subcaption}

\usepackage{booktabs}

\usepackage{hyperref}

\usepackage{url}

\usepackage{amsmath}

\usepackage{amssymb}

\usepackage{mathtools}

\usepackage{amsthm}

\usepackage[capitalize,noabbrev]{cleveref}

\graphicspath{{figures/}{icml2026/figures/}}

\icmltitlerunning{Representation in ML: Beyond Plato and Aristotle}

\begin{document}

\twocolumn[

\icmltitle{The Concept of Representation in ML: Beyond Plato and Aristotle}

\begin{icmlauthorlist}
\icmlauthor{Gilad D. Landau}{ox}

\icmlauthor{Aviv Keren}{aff2}

\end{icmlauthorlist}

\icmlaffiliation{ox}{University of Oxford}

\icmlaffiliation{aff2}{Sett AI}

\icmlcorrespondingauthor{Gilad D. Landau}{gilad.landau@jesus.ox.ac.uk}

\icmlkeywords{representation learning, philosophy of AI, machine learning theory}

\vskip 0.3in

]

\printAffiliationsAndNotice{}

\begin{abstract}
Representation is a central concept in modern machine learning, where it usually refers to internal encodings that support learning and generalization. As models scale and their capabilities become increasingly human-level, this representational language sometimes shifts from an engineering context into the more philosophically loaded domain of mental representation. We argue that this is the case for recent claims about the convergence of representational properties across different AI models. In particular, we assess the arguments developed in \emph{The Platonic Representation Hypothesis}, according to which this convergence is driven by a unified structure of reality. We examine this claim by introducing arguments and ideas from debates about mental representation in philosophy of mind. We argue that these philosophical resources can clarify what is at stake in such claims, explain why alignment evidence alone is insufficient for strong metaphysical conclusions, and suggest directions for future research.\end{abstract}

The term representation is widely used in machine learning. It is often invoked to describe a model's internal encoding of some aspect of its input, output, or task-relevant structure. Representational talk is especially useful for describing a model's ability to learn such encodings, to operate over them, to compress information efficiently, and to preserve desired properties. In this engineering-based context, ``representation'' functions as a flexible and productive term.

By contrast, in philosophy of mind and cognitive science, representation is a more contested concept \citep{pitt2024mentalrepresentation}. Philosophers have long distinguished genuine representation from mere correlation, covariance, or causal sensitivity. For a state of a system to count as a representation in a philosophically robust sense, it is usually not enough that it tracks some feature of the world. It must also play the right kind of role in the system, supporting content and other informational processes.

Until recently, the operational notion of representation used in machine learning was largely sufficient for engineering practice. But as models expand across modalities, scales, and tasks, and as increasingly strong claims are made about their cognitive capacities, the need for greater conceptual clarity becomes more pressing. One example is the recent discussion in machine learning around ``Unified Representations'' \cite{sucholutsky2024representationalalignment}: the main claim is that sufficiently large models, trained on different data, architectures, or modalities, converge on a shared representational structure. The strongest version of this claim is ``The Platonic Representation Hypothesis'' \citep{platonicHypothesis}, where the observation of representational convergence between different families of models receives a metaphysical interpretation: as models scale, their representations tend to converge because they increasingly approximate the structure of an underlying unified reality.

This paper argues that such claims move beyond the ordinary engineering usage of the term ``representation.'' The issue is not the use of representational language by machine-learning practitioners and researchers. In its technical sense, the term is well motivated and useful. Moreover, this usage is not wholly disconnected from the philosophical notion: in both contexts, representation involves a relation in which some state or structure is taken to stand for, encode, or carry information about some other aspect of the system or the world. In machine learning, however, this relation is usually invoked operationally and interpretively, without requiring a fuller account of what grounds it or fixes its content. The problem arises when representational convergence is used to support stronger claims about models representing the structure of reality. As we will show, such claims require an account of the foundational conceptual issues raised by philosophers of representation through many years of debate.

We draw on several philosophical resources to make this point. First, Fodor's disjunction problem shows why covariance or causal correlation alone cannot determine representational content \citep{fodor1987psychosemantics}. An internal state may covary with many distal objects, proximal causes, background conditions, and error-inducing cases. Correlation alone does not tell us which of these fixes the content of the representational state. Second, we review the teleosemantic response to this challenge and bring functional theories of representation to bear on the issue.

The central argument of this paper is therefore the following: current claims about the representational powers of large models are not yet accountable to the philosophically robust sense of representation. In particular, evidence of representational alignment between models does not by itself warrant conclusions about what those models represent, or whether they represent a unified structure of reality. A more robust account of representation would require additional work: individuating the relevant content-bearing vehicles within the system, explaining how they contribute to success and failure in task performance, and accounting for their causal role within the model's internal processing.

\section{Representation in ML}

The term representation is central to modern machine learning. Simply put, a representation is an internal encoding of data that a model can use to perform some operation. While a useful term for conceptualizing any form of learning algorithm, it took center stage with the rise of deep learning.

In one of the canonical papers that helped define deep learning as a field, the authors introduce deep learning in the following terms:
\begin{quote}
    Deep-learning methods are \textbf{representation-learning} methods with multiple levels of representation, obtained by composing simple but non-linear modules that each transform the representation at one level (starting with the raw input) into a representation at a higher, slightly more abstract level \citep[p.~436]{lecun2015deep}.
\end{quote}

Defining deep learning as a form of ``representation learning'' is therefore highly illustrative. Much of deep learning's success comes from its ability to learn useful internal representations from training signals, rather than relying on hand-crafted features. For the task of image classification, for example, a model trained end-to-end on raw pixels may develop early layers that function as edge detectors, capturing simple visual primitives, while later layers become sensitive to object parts and higher-level structures in the visual scene. Describing deep networks in these representational terms is one of the main ways to understand and interpret how they perform the tasks they were trained to do \citep{olah2018building}.

Representational language has remained central with the rise of self-supervised learning and foundation models. Large language models, vision-language models, and other generative systems are trained on internet-scale data using self-supervised objectives such as next-token prediction \citep{brown2020language}, masked-patch prediction \citep{he2022masked}, or contrastive alignment \citep{radford2021learning}. Their generality is often explained in representational terms: they are said to acquire rich, task-general internal representations that can be adapted to a wide range of downstream tasks \citep{devlin2019bert,he2022masked}.

The ``Unified Representations'' movement (``UniReps'', \citet{sucholutsky2024representationalalignment}) is a striking example of the growing interest in the representational capacities of large AI models. Its central claim is that, as models scale and become more capable, their internal representational spaces increasingly align across architectures, datasets, and modalities. At the limit of scale, they observe, all representational systems converge on similar representations. 

``The Platonic Representation Hypothesis'' \citep{platonicHypothesis}, a prominent paper associated with this movement, invokes the Platonic theory of Ideas \citep{plato1997republic} as an inspiration and builds on this observation. The authors hypothesize that the observed representational convergence is not driven by any technical properties of the models or their method of training, but by the fact that there is one underlying reality that any sufficiently capable model will learn to approximate.

The empirical basis for representational-alignment claims involves measuring the average kernel alignments of their internal activations. This method compares the relational geometry of activation spaces: Two representations \(f\) and \(g\) are representationally aligned if their induced kernels \(K_f\) and \(K_g\) are similar. 
That is, for inputs \(x_i, x_j \in \mathcal{X}\), the similarity assigned to the pair \((x_i,x_j)\) by \(K_f\) corresponds to the similarity assigned to the same pair by \(K_g\). 

\citet{platonicHypothesis} go on to show that as models get bigger and better (as evaluated on standard benchmarks), the kernel alignment measures grow. This phenomenon, they observe, persists  for models of different architectures and even across different modalities.

In the context of kernel alignment measurement, ``representation'' is being used in the ordinary engineering sense, referring to internal encodings of stimuli within a model's activation space. The shift to the philosophical notion of representation occurs when the observed alignment is used to support a metaphysical claim about the unified nature of reality. It is then that the hypothesis begins to carry a heavier philosophical burden. 
\section{Representation in Philosophy}

A central problem in philosophy of mind is how physical or computational states come to have content. The fact that the mind has states that are about things beyond themselves, whether real, possible, or imaginary, is what requires explanation. This is the issue of Mental Representation \citep{shea2018representation}.

Naively, we might think that a mental representation of a dog is a kind of brain activation that occurs whenever we see a dog, and so the content of the representation is fixed by the causal pathway from the actual dog to the corresponding brain activation. The philosopher Jerry Fodor argued that such causal determination is not enough to set representational content \citep{fodor1987psychosemantics}. There are many points along the causal pathway from the dog to the neural activation, such as the pattern of reflected light or the optical retinal projection, and it is unclear why one of these rather than another should determine content. Moreover, error cases create a deeper problem. If poor lighting or noise causes a horse to trigger the same activation, what makes that state represent a dog rather than a horse, or a dog-under-ideal-conditions? Fodor called this \textit{the disjunction problem}, and it is a deep and important issue that any good theory of representation needs to address \citep{fodor1987psychosemantics}.

One influential response to Fodor’s challenge comes from teleosemantic theories of mental representation \citep{millikan1984language}. Rather than identifying content with whatever happens to cause an internal state, these theories tie content to function. On classical versions of the view, associated especially with Millikan \citep{millikan1984language} and related informational theories such as Dretske's \citep{dretske1988explaining}, a state represents what it has the proper function of indicating, carrying information about, or enabling a system to respond to. This function is grounded in the evolutionary and developmental history of the system: the state has the content it does because systems of that kind evolved to use it in guiding successful behavior. Teleosemantic theories therefore offer a way to distinguish content-fixing relations from accidental correlations. What matters is not mere covariance, but how a state supports a proper function under the normal conditions in which that function was evolved and selected.

Teleosemantics faces several challenges, one of the most influential being Davidson's ``Swampman'' thought experiment \citep{davidson1987knowing}. Davidson asks us to imagine that lightning strikes a swamp and accidentally produces a molecule-for-molecule duplicate of a human being. Although this ``Swampman'' would be physically and functionally indistinguishable from an ordinary person, it would lack the historical background that classical teleosemantics treats as content-fixing. Its internal states would have the same causal organization as ours, but they would not have been shaped by evolutionary selection or by a history of successful use. The case therefore poses a challenge for evolutionary theories of representation: can a system with the right internal organization, but without the right history, possess genuine representational content?

This issue is especially relevant to artificial systems. Contemporary large language models are not biologically evolved organisms, even though they are highly organized systems that perform complex cognitive tasks. Any theory that grounds representational content solely in evolutionary history will therefore struggle to account for their representational capacities. The ``Swampman'' case thus helps isolate a central question for AI: whether representational content requires the right historical origin, or whether it can also arise from learning, design, stabilization, and functional role within a system.

Nicholas Shea's theory of representation offers a useful response to this challenge by relaxing many of the evolutionary commitments of classical teleosemantics \citep{shea2018representation}. In response to the ``Swampman'' case, Shea accepts that a newly created ``Swampman'' would not immediately have settled representational content. However, once such a system begins to interact with its environment, learn from feedback, and develop stabilized patterns of success and failure, its internal states can acquire representational content \citep[p.~22]{shea2018representation}. The functions relevant to representation therefore need not depend solely on biological evolution. They can also arise through learning, design, and stabilization, provided that internal states play reliable roles in guiding the system's successful behavior.

Before revisiting ``The Platonic Representation Hypothesis'' in light of these philosophical concerns, we can summarize the main points as follows. First, any theory of representation must contend with Fodor's disjunction problem: the problem of explaining how a representation can have determinate content despite also being causally connected to cases of error or misrepresentation. Second, if we want a causal theory of representational content, we need to consider the function that a representation serves within a larger system, as emphasized by teleosemantics. Third, the function of a representation is not fixed only by its causal origin, but also by the downstream subsystems that consume, interpret, and use it. In this sense, representational content depends not merely on what produces a state, but also on what the system is set up to do with that state.

\section{Revisiting the Platonic Representation Hypothesis}

Although the Platonic Representation Hypothesis is framed in terms of representational convergence toward a singular underlying reality, it can also be read more modestly: as the claim that different models may converge toward a shared statistical model of the structure that generates their observations. On this weaker reading, the metaphysical question concerning the nature of the underlying reality is left underdetermined. Nevertheless, \citet{platonicHypothesis} also situate the hypothesis in relation to ``Convergent Realism'' \citep{laudan1981confutation}, a realist view in the philosophy of science according to which successful scientific theories progressively disclose the structure of the world \citep{putnam1982threekinds}. If this framing is taken seriously, then it is hard to avoid the stronger reading according to which convergence is not merely evidence of shared statistical structure, but evidence that models are increasingly recovering the structure of reality itself. Our concern is with this latter move: from the observation of representational convergence to a metaphysical view of a single underlying reality.

Our argument is that, in order to sustain this kind of claim, the use of representational language must be able to withstand the philosophical challenges raised above. It is not enough to show that different models converge on similar internal structures. One must also explain in what sense these structures are representations, what fixes their content, what functions they serve within the system, and how they avoid collapsing into mere correlations or useful engineering abstractions.

The Platonic Representation Hypothesis, as it stands, does not yet meet these criteria. In particular, its main analysis and empirical support rely heavily on kernel alignment methods, which measure structural or covariance-based similarities between stimuli and activation patterns across different layers of models. These methods may show that different systems organize information in similar ways, but they do not by themselves provide a criterion for fixing representational content.

As such, covariance-based methods such as those used to validate the hypothesis cannot resolve the disjunction problem. They can show that two systems exhibit similar relational structures, but they cannot explain why a given internal state represents one content rather than another, or how the system distinguishes veridical representation from error, noise, or merely correlated structure.

This is, admittedly, a philosophical issue rather than a purely technical one. However, if ``The Platonic Representation Hypothesis'' aims to make a philosophical claim about representation and reality, especially one with strong metaphysical implications, then it must either meet this challenge or explicitly engage with it. Without such an account, the evidence for representational convergence remains compatible with a much weaker claim: that models trained on large datasets develop similar useful structures, not that they are converging on reality itself.

Interestingly, a recent follow-up paper, ``Revisiting the Platonic Representation Hypothesis: An Aristotelian View'' \citep{groeger2026aristotelian}, challenges the Platonic Representation Hypothesis on its own technical grounds, suggesting that the disjunction problem may have concrete practical consequences. 

The authors prove mathematically that, as models scale, the number of possible spurious correlations between model activations and target properties also grows. This makes it increasingly unclear which activations genuinely contribute to a model's representational content and which merely correlate with the property under investigation \citep{groeger2026aristotelian}.

On this view, the apparent alignment reported by the Platonic Representation Hypothesis may partly reflect the expanding space of possible random or spurious correlations, rather than genuine convergence on a shared underlying reality. When the authors controlled for this growth in activation space, many of the previously reported alignment effects disappeared. Some more modest forms of alignment remained, which they interpret as supporting an ``Aristotelian'' rather than ``Platonic'' view: models may converge on local, task-relevant structures, but this weaker form of convergence does not warrant further metaphysical claims about a unified structure of reality \citep{groeger2026aristotelian}.

This result is especially relevant to our argument because it shows that the philosophical worry is not merely abstract. If representational content is inferred only from covariance or structural similarity, then larger models may appear to be more aligned simply because they provide more opportunities for accidental alignment. Thus, without a stronger criterion for content-fixing, the Platonic Representation Hypothesis risks mistaking spurious correlation for genuine representation.

On the positive side, the philosophy of mental representation also offers useful guidance for future analysis. If large language models are to be treated as genuine representational systems, then it is crucial to understand how their representations operate as part of the functioning of the system as a whole. Although large language models are not evolved biological systems, Shea's response to the ``Swampman'' thought experiment suggests that this need not be a decisive obstacle. On Shea's view, representational capacities can be acquired through interaction with the environment, learning, and the stabilization of successful patterns of use.

Since function is central to the fixing of representational content, this raises a further concern: whether it is coherent to assess representational content independently of any specific task. A stronger future version of the Platonic Representation Hypothesis may therefore require a more active, task-based analysis of representation. Rather than asking whether model activations converge in the abstract, we may need to ask what role those activations play in enabling the model to perform particular functions.

Finally, Shea's emphasis on the downstream consumers of representations suggests that mechanistic analysis of large language models is essential for understanding what, and how, they represent. Work on mechanistic interpretability in deep learning \cite{olah2020zoom} aims to provide causal and mechanistic explanations of how neural networks operate. Such analyses can identify which components of a model interact with other components in order to carry out a target task. In principle, this kind of analysis could help individuate the relevant representational vehicles and clarify the content they carry \citep{cunningham2023sparse}.

\section{Conclusion}

We have argued that the operational concept of representation in machine learning is both useful and important for engineering practice. It allows researchers to describe, compare, and manipulate internal model states in technically productive ways. Nevertheless, as models become larger and acquire more cognitively impressive capacities, there is a growing temptation to move from this operational usage of representation to a more philosophical concept of representation. This transition is legitimate, but it brings with it a philosophical burden: one must explain what the representational relation consists in, and how representational content is fixed.

Fodor's disjunction problem brings this challenge sharply into view. Any serious theory of representation must explain how a state can have determinate content, rather than merely standing in a pattern of causal or statistical relations to many possible objects, properties, or environmental conditions. We have suggested that teleosemantic theories offer one way of understanding what such a content-fixing procedure might require. In particular, they emphasize the functional role of a representation within a larger system, and the importance of downstream consumers that use the represented content to guide further processing or action.

Philosophical analysis can do more than criticize grand metaphysical claims in machine learning. It can clarify the conditions under which stronger representational claims would be justified, and it can guide concrete research practices for evaluating what large models represent and how they represent it.

\bibliographystyle{icml2026}
\bibliography{paper}

@book{fodor1987psychosemantics,
  author    = {Jerry A. Fodor},
  title     = {Psychosemantics: The Problem of Meaning in the Philosophy of Mind},
  publisher = {MIT Press},
  address   = {Cambridge, MA},
  year      = {1987}
}

@book{shea2018representation,
  author    = {Nicholas Shea},
  title     = {Representation in Cognitive Science},
  publisher = {Oxford University Press},
  address   = {Oxford, UK},
  year      = {2018}
}

@article{lecun2015deep,
  author  = {Yann LeCun and Yoshua Bengio and Geoffrey Hinton},
  title   = {Deep Learning},
  journal = {Nature},
  volume  = {521},
  number  = {7553},
  pages   = {436--444},
  year    = {2015}
}

@article{olah2018building,
  author  = {Chris Olah and Nick Cammarata and Ludwig Schubert and Gabriel Goh and Michael Petrov and Shan Carter},
  title   = {The Building Blocks of Interpretability},
  journal = {Distill},
  volume  = {3},
  number  = {3},
  pages   = {e10},
  year    = {2018},
  doi     = {10.23915/distill.00010}
}

@misc{platonicHypothesis,
  author        = {Minyoung Huh and Brian Cheung and Tongzhou Wang and Phillip Isola},
  title         = {The Platonic Representation Hypothesis},
  year          = {2024},
  eprint        = {2405.07987},
  archivePrefix = {arXiv},
  primaryClass  = {cs.LG},
  url           = {https://arxiv.org/abs/2405.07987}
}

@misc{groeger2026aristotelian,
  author        = {Fabian Gr{\"o}ger and Shuo Wen and Maria Brbi{\'c}},
  title         = {Revisiting the Platonic Representation Hypothesis: An Aristotelian View},
  year          = {2026},
  eprint        = {2602.14486},
  archivePrefix = {arXiv},
  primaryClass  = {cs.LG},
  url           = {https://arxiv.org/abs/2602.14486}
}

@book{millikan1984language,
  author    = {Ruth Garrett Millikan},
  title     = {Language, Thought, and Other Biological Categories: New Foundations for Realism},
  publisher = {MIT Press},
  address   = {Cambridge, MA},
  year      = {1984}
}

@book{dretske1988explaining,
  author    = {Fred Dretske},
  title     = {Explaining Behavior: Reasons in a World of Causes},
  publisher = {MIT Press},
  address   = {Cambridge, MA},
  year      = {1988}
}

@incollection{davidson1987knowing,
  author    = {Donald Davidson},
  title     = {Knowing One's Own Mind},
  booktitle = {Proceedings and Addresses of the American Philosophical Association},
  volume    = {60},
  number    = {3},
  pages     = {441--458},
  year      = {1987}
}

@article{sucholutsky2024representationalalignment,
  title={Getting aligned on representational alignment},
  author={Sucholutsky, Ilia and Muttenthaler, Lukas and Weller, Adrian and Peng, Andi and Bobu, Andreea and Kim, Been and Love, Bradley C. and Cueva, Christopher J. and Grant, Erin and Groen, Iris and Achterberg, Jascha and Tenenbaum, Joshua B. and Collins, Katherine M. and Hermann, Katherine L. and Oktar, Kerem and Greff, Klaus and Hebart, Martin N. and Cloos, Nathan and Kriegeskorte, Nikolaus and Jacoby, Nori and Zhang, Qiuyi and Marjieh, Raja and Geirhos, Robert and Chen, Sherol and Kornblith, Simon and Rane, Sunayana and Konkle, Talia and O'Connell, Thomas P. and Unterthiner, Thomas and Lampinen, Andrew K. and M{\"u}ller, Klaus-Robert and Toneva, Mariya and Griffiths, Thomas L.},
  journal={arXiv preprint arXiv:2310.13018},
  year={2024},
  doi={10.48550/arXiv.2310.13018},
  url={https://arxiv.org/abs/2310.13018}
}

@inproceedings{devlin2019bert,
title={{BERT}: Pre-training of Deep Bidirectional Transformers for Language Understanding},
author={Devlin, Jacob and Chang, Ming-Wei and Lee, Kenton and Toutanova, Kristina},
booktitle={Proceedings of {NAACL-HLT}},
pages={4171–4186},
year={2019}
}

@inproceedings{he2022masked,

  title={Masked Autoencoders Are Scalable Vision Learners},

  author={He, Kaiming and Chen, Xinlei and Xie, Saining and Li, Yanghao and Doll{\'a}r, Piotr and Girshick, Ross},

  booktitle={Proceedings of the IEEE/CVF Conference on Computer Vision and Pattern Recognition},

  pages={16000--16009},

  year={2022}

}

@article{olah2020zoom,

  title={Zoom In: An Introduction to Circuits},

  author={Olah, Chris and Cammarata, Nick and Schubert, Ludwig and Goh, Gabriel and Petrov, Michael and Carter, Shan},

  journal={Distill},

  year={2020},

  url={https://distill.pub/2020/circuits/zoom-in}

}

@article{cunningham2023sparse,

  title={Sparse Autoencoders Find Highly Interpretable Features in Language Models},

  author={Cunningham, Hoagy and Ewart, Aidan and Riggs, Logan and Huben, Robert and Sharkey, Lee},

  journal={arXiv preprint arXiv:2309.08600},

  year={2023},

  doi={10.48550/arXiv.2309.08600},

  url={https://arxiv.org/abs/2309.08600}

}

@incollection{pitt2024mentalrepresentation,

  title={Mental Representation},

  author={Pitt, David},

  booktitle={The Stanford Encyclopedia of Philosophy},

  editor={Zalta, Edward N. and Nodelman, Uri},

  year={2024},

  publisher={Metaphysics Research Lab, Stanford University},

  url={https://plato.stanford.edu/entries/mental-representation/}

}

@inproceedings{brown2020language,

  title={Language Models are Few-Shot Learners},

  author={Brown, Tom B. and Mann, Benjamin and Ryder, Nick and Subbiah, Melanie and Kaplan, Jared and Dhariwal, Prafulla and Neelakantan, Arvind and Shyam, Pranav and Sastry, Girish and Askell, Amanda and Agarwal, Sandhini and Herbert-Voss, Ariel and Krueger, Gretchen and Henighan, Tom and Child, Rewon and Ramesh, Aditya and Ziegler, Daniel M. and Wu, Jeffrey and Winter, Clemens and Hesse, Christopher and Chen, Mark and Sigler, Eric and Litwin, Mateusz and Gray, Scott and Chess, Benjamin and Clark, Jack and Berner, Christopher and McCandlish, Sam and Radford, Alec and Sutskever, Ilya and Amodei, Dario},

  booktitle={Advances in Neural Information Processing Systems},

  volume={33},

  pages={1877--1901},

  year={2020}

}

@inproceedings{radford2021learning,

  title={Learning Transferable Visual Models From Natural Language Supervision},

  author={Radford, Alec and Kim, Jong Wook and Hallacy, Chris and Ramesh, Aditya and Goh, Gabriel and Agarwal, Sandhini and Sastry, Girish and Askell, Amanda and Mishkin, Pamela and Clark, Jack and Krueger, Gretchen and Sutskever, Ilya},

  booktitle={Proceedings of the 38th International Conference on Machine Learning},

  pages={8748--8763},

  year={2021},

  publisher={PMLR}

}

@incollection{plato1997republic,
  title={Republic},
  author={Plato},
  booktitle={Plato: Complete Works},
  editor={Cooper, John M.},
  translator={Grube, G. M. A. and Reeve, C. D. C.},
  publisher={Hackett Publishing},
  address={Indianapolis, IN},
  year={1997},
  pages={971--1223}
}

@article{putnam1982threekinds,
  author  = {Putnam, Hilary},
  title   = {Three Kinds of Scientific Realism},
  journal = {The Philosophical Quarterly},
  volume  = {32},
  number  = {128},
  pages   = {195--200},
  year    = {1982}
}

@article{laudan1981confutation,
  author  = {Laudan, Larry},
  title   = {A Confutation of Convergent Realism},
  journal = {Philosophy of Science},
  volume  = {48},
  number  = {1},
  pages   = {19--49},
  year    = {1981},
  doi     = {10.1086/288975}
}

\end{document}